\documentclass[lettersize,journal]{IEEEtran}
\usepackage{amsmath,amsfonts}
\usepackage{algorithmic}
\usepackage{algorithm}
\usepackage{array}

\usepackage{booktabs}

\usepackage{cite}
\usepackage[caption=false,font=normalsize,labelfont=sf,textfont=sf]{subfig}
\usepackage{textcomp}
\usepackage{stfloats}
\usepackage{url}
\usepackage{verbatim}
\usepackage{graphicx}
\usepackage{cite}
\begin{document}

\title{SHARP-Net: A Refined Pyramid Network for Deficiency Segmentation in Culverts and Sewer Pipes}

\author{IEEE Publication Technology,~\IEEEmembership{Staff,~IEEE,}

\author{
    \IEEEauthorblockN{Rasha Alshawi\IEEEauthorrefmark{1}, Md Meftahul Ferdaus\IEEEauthorrefmark{1}, Md Tamjidul Hoque\IEEEauthorrefmark{1}, Kendall Niles\IEEEauthorrefmark{2}, Ken Pathak\IEEEauthorrefmark{2}, Steve Sloan\IEEEauthorrefmark{2}, Mahdi Abdelguerfi\IEEEauthorrefmark{1}}
    \IEEEauthorblockA{\IEEEauthorrefmark{1}University of New Orleans, New Orleans, Louisiana, USA}
    \IEEEauthorblockA{\IEEEauthorrefmark{2}US Army Corps of Engineers, Vicksburg, Mississippi, USA}
    \IEEEauthorblockA{Emails: \{rralshaw, mferdaus,thoque\}@uno.edu, \{Kendall.N.Niles, ken.pathak, steven.d.sloan\}@erdc.dren.mil, GulfsceiDirector@uno.edu}
}

\thanks{}
\thanks{Manuscript received April 19, 2021; revised August 16, 2021.}}

\markboth{Journal of \LaTeX\ Class Files,~Vol.~14, No.~8, August~2021}%
{Shell \MakeLowercase{\textit{et al.}}: A Sample Article Using IEEEtran.cls for IEEE Journals}



\maketitle

\begin{abstract}

This paper introduces Semantic Haar-Adaptive Refined Pyramid Network (SHARP-Net), a novel architecture for semantic segmentation. SHARP-Net integrates a bottom-up pathway featuring Inception-like blocks with varying filter sizes ($3\times3$ and $5\times5$), parallel max-pooling, and additional spatial detection layers. This design captures multi-scale features and fine structural details. Throughout the network, depth-wise separable convolutions are used to reduce complexity. The top-down pathway of SHARP-Net focuses on generating high-resolution features through upsampling and information fusion using $1\times1$ and $3\times3$ depth-wise separable convolutions. We evaluated our model using our developed challenging Culvert-Sewer Defects dataset and the benchmark DeepGlobe Land Cover dataset. Our experimental evaluation demonstrated the base model's (excluding Haar-like features) effectiveness in handling irregular defect shapes, occlusions, and class imbalances. It outperformed state-of-the-art methods, including U-Net, CBAM U-Net, ASCU-Net, FPN, and SegFormer, achieving average improvements of 14.4\% and 12.1\% on the Culvert-Sewer Defects and DeepGlobe Land Cover datasets, respectively, with IoU scores of 77.2\% and 70.6\%. Additionally, the training time was reduced. Furthermore, the integration of carefully selected and fine-tuned Haar-like features enhanced the performance of deep learning models by at least 20\%. The proposed SHARP-Net, incorporating Haar-like features, achieved an impressive IoU of 94.75\%, representing a 22.74\% improvement over the base model. These features were also applied to other deep learning models, showing a 35.0\% improvement, proving their versatility and effectiveness. SHARP-Net thus provides a powerful and efficient solution for accurate semantic segmentation in challenging real-world scenarios.

\end{abstract}


\section{Introduction}

\IEEEPARstart{A}{ccurate} detection and segmentation of defects in culverts and sewer pipes is crucial for effective infrastructure management, playing a vital role in ensuring the safety and integrity of underground utilities \cite{abdel2010risk}. Undetected defects can lead to severe consequences, including structural failures, increased maintenance costs, and environmental hazards. Therefore, automating and enhancing defect detection through advanced computer vision techniques presents significant opportunities for improving infrastructure management and safety \cite{arnoult1986culvert}.

Traditional defect detection methods involve manual inspection and assessment, which is time-consuming and prone to human error. Advanced computer vision techniques, like semantic segmentation, offer potential to automate these processes \cite{iqbal2022prediction}. Semantic segmentation assigns pixel-level labels to objects or regions in an image, making it a powerful tool for understanding and analyzing visual scenes \cite{abdelrahman2022kidney, ren2023visual}. This technique is well-suited for culvert and sewer systems, enabling more accurate defect detection, assessment, and maintenance planning.

Despite its potential, applying semantic segmentation to culvert and sewer pipe inspection presents several challenges \cite{guo2009automated, grier2022large}. The visual characteristics of these environments are highly diverse, with variations in scale, orientation, appearance, and environmental conditions such as occlusions and lighting changes \cite{alshawi2016understanding}. Moreover, the datasets available for training models in this domain are often limited and imbalanced, making it challenging to achieve high performance with standard segmentation approaches \cite{gao2021use}.

Current solutions for defect detection in these contexts often fall short due to their inability to handle the full complexity of real-world environments. Traditional models may not adequately address variations in defect types, pipe materials, and environmental conditions. In response to these challenges, we propose SHARP-Net (Semantic Haar-Adaptive Refined Pyramid Network), an innovative approach designed to tackle the complexities inherent in semantic segmentation tasks involving culverts and sewer pipes. SHARP-Net combines hierarchical feature representations extracted by Feature Pyramid Networks (FPN) with advanced enhancements in feature extraction to improve object segmentation and localization accuracy. By integrating multi-scale feature maps, sparsely connected blocks, and fine-tuned Haar-like features, SHARP-Net aims to achieve superior performance in accurately detecting defects while maintaining computational efficiency (in terms of number of parameters).

In addition to its performance on our dataset, we evaluated SHARP-Net on another benchmark dataset to assess its generalizability across different contexts. This evaluation demonstrated that SHARP-Net maintains its effectiveness and robustness, achieving competitive results on diverse datasets and confirming its capability to handle a wide range of semantic segmentation tasks beyond the specific culvert and sewer defect domain.

The main contributions of this paper are as follows: 
\begin{itemize}
    \item We present SHARP-Net, an innovative architecture specifically designed for semantic segmentation of defects in culverts and sewer pipes. SHARP-Net incorporates Inception-like blocks, depth-wise separable convolutions, and a top-down pathway that generates high-resolution features by upsampling and fusing information.
    
    \item To improve SHARP-Net's performance, we incorporate fine-tuned Haar-like features that capture critical edge, line, and corner information necessary for distinguishing defect classes in the challenging culvert and sewer pipe dataset.
    
\end{itemize}

We demonstrate that SHARP-Net outperforms state-of-the-art methods through experiments on our Culvert-Sewer Defects dataset and benchmark DeepGlobe Land Cover dataset, setting a new standard for accuracy and efficiency in semantic segmentation. The code for our proposed models is publicly available at: https://github.com/RashaAlshawi/HFFPN.

\section{Related Work}

Various architectures have been developed for semantic segmentation, with prominent approaches including bottom-up top-down networks like FPNs and encoder-decoder networks (EDNs) like U-Net \cite{lin2017feature, ronneberger2015u, oluwasammi2021features}. FPNs efficiently address multi-scale feature extraction by constructing a hierarchical pyramid of feature maps at different resolutions, integrating contextual information to enhance robustness and accuracy. Lin et al. \cite{lin2017feature} demonstrated the efficacy of FPNs in object detection, utilizing a ResNet backbone trained on ImageNet \cite{deng2009imagenet} to extract hierarchical features through bottom-up and top-down pathways.

Conversely, EDNs like U-Net efficiently capture spatial dependencies and preserve high-resolution features through skip connections, which makes them highly effective for precise object localization \cite{alshawi2023depth}. Variants such as the Convolutional Block Attention Module (CBAM) \cite{su2022research} and Attention Sparse Convolutional U-Net (ASCU-Net) \cite{tong2021ascu} further enhance U-Net's performance. CBAM improves U-Net by adding attention mechanisms that refine feature extraction in two stages: channel attention emphasizes important feature channels, and spatial attention focuses on relevant regions within the feature maps, resulting in more precise segmentation. ASCU-Net integrates attention mechanisms with sparse convolutional layers to handle irregular and sparse features more effectively. Its attention module dynamically prioritizes important features while the sparse convolutions reduce computational complexity, thereby enhancing the model's efficiency with complex and varied datasets.

Vision Transformers (ViTs) have become a leading method in computer vision, especially for tasks like image classification and object detection, due to their use of self-attention mechanisms that capture global dependencies across image patches \cite{dosovitskiy2020image}. SegFormer \cite{xie2021segformer} adapts this transformer architecture for semantic segmentation. It employs a transformer encoder to capture global context and relationships within the image, overcoming the limitations of traditional convolutional methods. During the decoding phase, SegFormer uses dense layers to create detailed pixel-level segmentation masks. This approach leverages transformers' ability to maintain global context while ensuring accurate spatial representation, making it effective for complex segmentation tasks.


Encoder-Decoder Networks (EDNs) offer precise localization but struggle with varied object scales and complex spatial arrangements \cite{yuan2023rmau, liu2019net}. FPNs handle multi-scale objects well but may not effectively address class imbalance \cite{gao2018end}. Vision Transformers (ViTs) capture global dependencies but can be computationally intensive and may miss fine-grained details. Applying these existing architectures directly to our diverse culvert-sewer defect dataset may be suboptimal, necessitating a tailored approach to address the specific challenges of varied defect types, sizes, and shapes.

Recent work in sewer and culvert inspection using deep learning has highlighted some specialized approaches to these specific challenges. For instance, several studies have adapted convolutional neural networks (CNNs) and EDNs to detect and classify defects in sewer systems. Zhang et al. \cite{zhang2023automatic} developed a deep learning framework that utilizes multi-scale feature extraction and data augmentation to address the issue of imbalanced defect types in sewer inspections. Similarly, Lee et al. \cite{lee2021automated} proposed an automated system that integrates CNNs with domain-specific pre-processing techniques to enhance defect detection accuracy in culvert inspections. Despite these advancements, current methods struggle to handle the diverse and complex nature of defects across varied environmental conditions. 

This paper presents a new architecture designed to address the challenges identified in current semantic segmentation approaches. We propose a technique to improve deep learning models and speed up their training.We tested our model's effectiveness using our dataset for segmenting culvert and sewer pipe defects, and a benchmark dataset to evaluate its versatility. Section \ref{sec:architecture} provides a comprehensive overview of the model's structure.

%
%
\section {SHARP-Net: Semantic Haar-Adaptive Refined Pyramid Network} \label{sec:architecture}

This section is divided into three subsections: Section \ref{sec:basemodel} discusses the architecture of SHARP-Net's base model, excluding Haar-like features. Section \ref{sec:ModelAnalysis} provides a comprehensive analysis of the architectural innovations and ablation studies conducted to develop the proposed model. It includes expanded results and insights into the various modifications explored. Section \ref{sec:Haar-inclusion} focuses on the integration of Haar-like features, detailing their extraction process and incorporation into the SHARP-Net model.

\subsection{SHARP-Net Base Architecture} \label{sec:basemodel}

The proposed model represents a significant advancement over the original FPN by incorporating an enhanced inception-like block within the bottom-up pathway. This addition improves the model’s ability to learn diverse and fine-grained features essential for accurate image analysis. Additionally, the use of depth-wise separable convolutions reduces model complexity while enhancing its ability to capture detailed information effectively.

The architecture is structured around two pathways, each playing a crucial role in feature refinement:

\begin{itemize}
\item \textbf{Bottom-Up Pathway:} This pathway utilizes inception-like blocks to enhance the model's ability to localize and detect objects in input images. These blocks process feature maps using a combination of filters with varying sizes ($3 \times 3$ and $5 \times 5$) and parallel max-pooling layers. Multiple filter sizes capture a wide range of spatial information for objects of different scales. Specifically, $3 \times 3$ filters capture fine details and textures crucial for detecting smaller objects or subtle features. Conversely, $5 \times 5$ filters capture broader features essential for recognizing larger objects or structures. Parallel max-pooling layers help the model retain spatial hierarchies, enhancing robustness to object position variations.

Depth-wise separable convolutions (depth-wise followed by point-wise convolutions) improve the model’s efficiency. This approach reduces parameters and computational complexity without compromising performance. This decomposition enhances computational efficiency and allows for a more flexible and fine-grained analysis of input features. The depth-wise convolution applies a single filter to each input channel separately, capturing spatial features while maintaining channel independence. The point-wise convolution combines the outputs of the depth-wise convolution by applying a $1 \times 1$ convolution, effectively mixing information across different channels. This approach reduces parameters and computational complexity without compromising performance.

Max-pooling with a stride of 2 is used to manage the spatial dimensions of the feature maps. This operation reduces the spatial resolution of the feature maps as they pass through the network.

\item \textbf{Top-Down Pathway:} The top-down pathway complements the bottom-up pathway. It generates higher-resolution features through upsampling operations and feature fusion. It starts with a $1 \times 1$ convolution to reduce the channel depth of the feature maps to 128, aligning it with the depth of the final bottom-up layer. This reduction maintains consistency between the feature maps from both pathways, facilitating integration during the fusion process.

Each subsequent layer in the top-down pathway is upsampled by a factor of 2, which increases the spatial resolution of the feature maps. After upsampling, the higher-resolution features are merged with the corresponding feature maps from the bottom-up pathway using a $1 \times 1$ convolution. This combines the refined top-down features with the contextually rich, lower-resolution features, ensuring alignment in channel depth for seamless integration.

To address aliasing effects during merging and preserve fine details, a $3 \times 3$ depth-wise separable convolution is applied. This layer helps maintain sharp transitions and complex details in the feature maps.

\item \textbf{Common Classifier:} A shared classifier across all output feature maps ensures consistency with a 128-dimensional output channel configuration. This facilitates efficient decision-making across diverse image contexts while optimizing computational resources.

 \end{itemize}

\begin{figure*}[ht]
    \centering
    \includegraphics[width=0.99\linewidth]{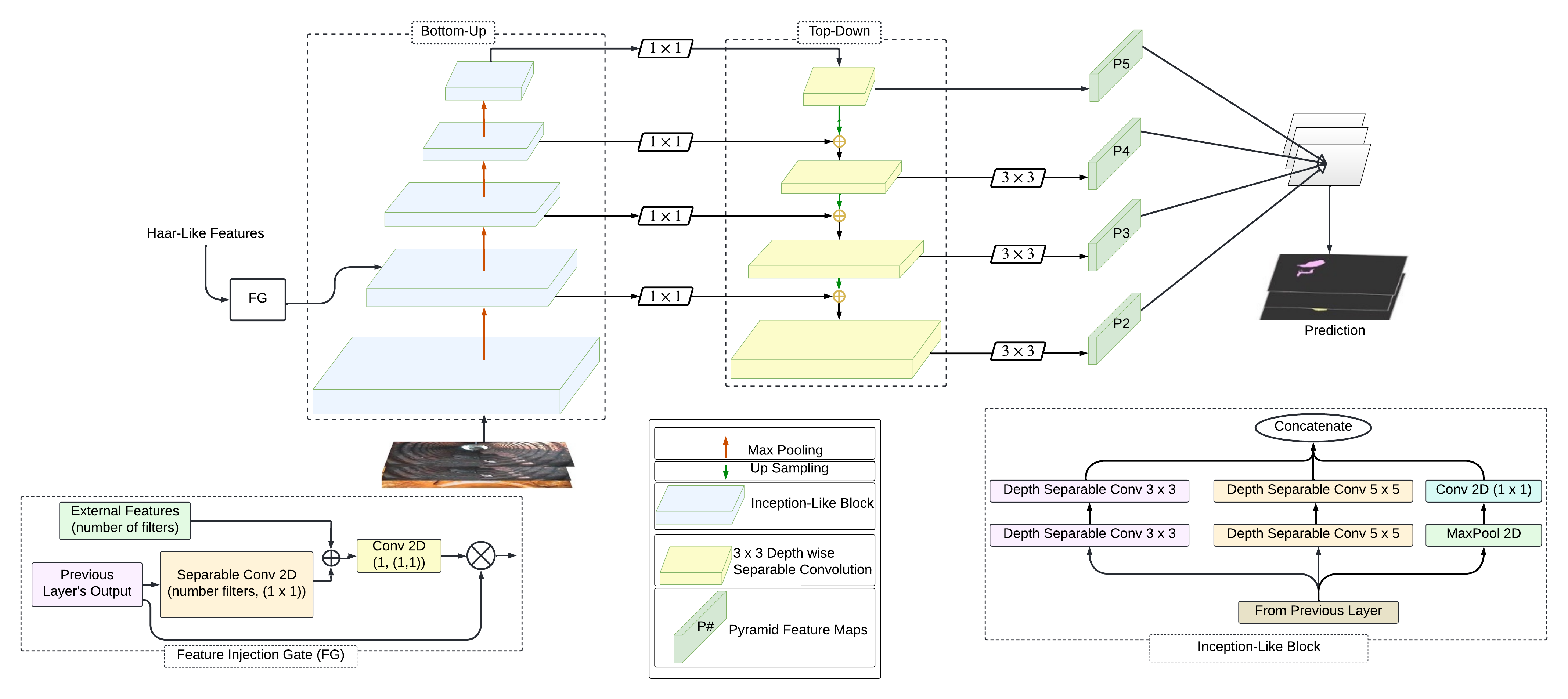}
  \caption{Architecture of the proposed SHARP-Net. The input image is progressively filtered and down-sampled by a factor of 2 at each layer in the bottom-up pathway on the left. The top-down pathway on the right performs up-sampling operations to reconstruct a colored-masked image. Haar-like features are injected into the second layer of the bottom-up pathway using feature injection gate shown on the bottom left of the figure.}\label{fig:model}

    \label{fig:EFPN}
\end{figure*}

The design of the Bottom-Up Pathway efficiently detects and localizes objects of varying sizes in input images. It uses inception-like blocks, depth-wise separable convolutions, and max-pooling operations. The top-down pathway refines and enhances the spatial resolution of the features, ensuring detailed and accurate output. Figure \ref{fig:EFPN} visually illustrates the architecture, highlighting the strategic integration of diverse filters and efficient feature handling.
\setlength{\parskip}{1em} 


\subsection{Architectural Evolution: From FPN to SHARP-Net}\label{sec:ModelAnalysis}

SHARP-Net evolved from extensive testing, incorporating key advanced elements into the FPN framework. This section details architectural enhancements to the original FPN, focusing on improving semantic segmentation performance. We aimed to find and apply the best methods to improve model accuracy and performance. Here are the key architectural changes from FPN to SHARP-Net:

\begin {itemize}
\item \textbf{Inception Block and Residual Connections:} We enhanced the FPN's Bottom-Up pathway by integrating Inception blocks and residual connections. This modification improves multi-scale feature extraction by allowing simultaneous processing of information through multiple filter sizes, capturing features at various scales. Residual connections improve deep network training by easing gradient flow and reducing vanishing gradients. This enhancement to the original FPN increased the IoU score to 0.74932, signifying better feature extraction and representation.

\item \textbf{Factorized Inception Block:} We improved computational efficiency by using a factorized Inception block, which breaks down large convolutions into smaller operations like $1 \times 1$ and $3 \times 3$ convolutions. This approach reduces computational demands and model size while preserving performance. Although slightly less effective than the full Inception block, the factorized version still outperformed the original FPN, achieving an IoU of 0.71863. This result highlights the trade-off between efficiency and performance.

\item \textbf{FPN with Atrous Convolutions:}
We integrated atrous convolutions into the FPN framework to expand the model's receptive field without increasing parameters or sacrificing spatial resolution. This aimed to enhance the capture of contextual information crucial for semantic segmentation, improving multi-scale feature extraction and preserving fine-grained details. Atrous convolutions achieve this by inserting spaces between kernel elements, allowing for larger receptive fields in a single operation. However, this approach presented challenges, including increased computational complexity, potential feature sparsity with large dilation rates, and grid effects in the output. To mitigate these issues, we experimented with various dilation rates and hybrid approaches combining standard and atrous convolutions. Despite the theoretical advantages and our efforts to optimize their implementation, the incorporation of atrous convolutions did not yield significant performance improvements. This underscores the complexity of architectural design in deep learning and the importance of empirical validation.

\item \textbf{FPN with Self-Attention Mechanisms:}
Self-attention, a key feature of Transformer models, allows the system to prioritize relevant parts of input sequences, capturing long-range dependencies and global context. It dynamically computes weighted representations, focusing on important information while ignoring less relevant parts. However, when integrated with atrous convolutions in the FPN model, this approach yielded a lower IoU of 0.644. This suggests that for this specific dataset, self-attention's ability to capture global dependencies did not significantly improve model performance, possibly due to challenges in combining self-attention with FPN or the dataset's unique characteristics.

\item \textbf{FPN with Attention Gates and Squeeze-and-Excitation Blocks:}
This configuration enhances FPN with Attention Gates and Squeeze-and-Excitation (SE) Blocks. Attention Gates dynamically highlight crucial regions in feature maps, focusing the network on relevant information. SE Blocks recalibrate channel-wise feature responses, capturing interdependencies between channels and improving feature representation. The combination of these techniques resulted in an improved IoU score of 0.75914, demonstrating enhanced accuracy and robustness in semantic segmentation tasks. This integration effectively prioritizes important features while suppressing noise, leading to better overall performance.

\end{itemize}

SHARP-Net emerged as the result of our FPN modifications. The key innovation, an Inception-like block with depth-wise separable convolutions, significantly improved accuracy and robustness while maintaining computational efficiency. This approach optimally balances model complexity and performance, addressing semantic segmentation challenges in complex infrastructure imagery. SHARP-Net's architecture enhances fine-grained detail capture and global context understanding, advancing semantic segmentation for infrastructure analysis and related fields.

\subsection{Haar-Like Feature Injection} \label{sec:Haar-inclusion}
To improve SHARP-Net's performance, we incorporated Haar-like features extracted from our dataset. While deep learning models often reduce the need for manual feature engineering, domain-specific features can be beneficial, especially with limited data, class imbalance, or few classes, as in our ten-class dataset \cite{alshawi2023dual}. Haar-like features, consisting of simple rectangular patterns, are effective for edge detection, line identification, and texture analysis. These computationally efficient features complement SHARP-Net's deep learning capabilities, potentially addressing challenges in defect segmentation for culvert and sewer pipe imagery.

\begin{figure}
    \centering
    \includegraphics[width=0.99\linewidth]{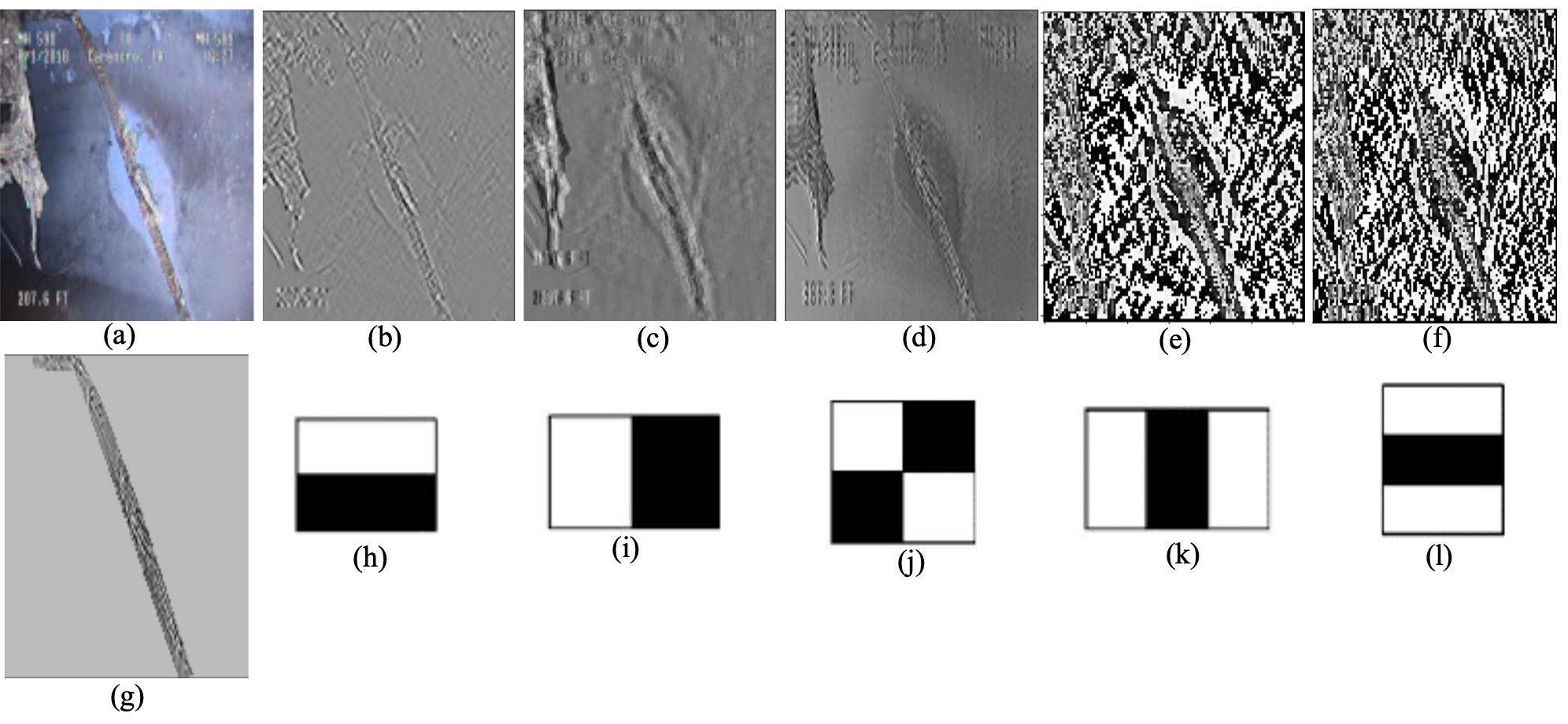}
    \caption{Applying Haar-like filters for feature extraction: (a) Original image, (b-f) Filter responses from edge and line detection filters, each extracted using the corresponding Haar-like filters shown below them (h-l). (g) An example of the filter response after applying noise reduction using mask region-based method.}
    \label{fig:haarfilter}
\end{figure}

Our Haar-like feature implementation for culvert and sewer pipe imagery focused on three key aspects: 1. Line detection: We used vertically elongated rectangles to capture the predominant vertical structures. 2. Edge detection: Symmetric windows (squares or similar-sized rectangles) were employed to identify sudden intensity or color changes at object boundaries. 3. Diagonal detection: A diagonal line detector was added to identify defects with both horizontal and vertical components. This comprehensive approach, illustrated in Figure \ref{fig:haarfilter} (second row), enhances SHARP-Net's ability to detect various defects \cite{maryan2019machine, panta2023iterlunet}.


Haar-like features and cascade classifiers perform best with power-of-2 window sizes, as shown by Viola and Jones \cite{viola2001rapid}. We tested various power-of-2 window sizes and Haar feature types on our dataset, using Peak Signal-to-Noise Ratio (PSNR) to assess image quality and feature detection accuracy \cite{korhonen2012peak}. Our analysis revealed that larger window sizes generally yielded higher PSNR scores, indicating better detection of sharp features crucial for defect identification. We found optimal window sizes of (4,2), (4,4), (8,4), and (16,4), balancing detection accuracy and computational efficiency. For detailed performance metrics and analyses, see Appendix \ref{sec:HaarAnalysis}.  


After determining optimal window sizes, we conducted a feature selection process to ensure high-quality, diverse Haar-like features. We used Peak Signal-to-Noise Ratio (PSNR) to measure image similarity, with values above 20 indicating high similarity. To maintain diversity and reduce redundancy, we retained features with distinct PSNR values while excluding those with PSNR values of 18 or higher, as shown in Table \ref{tab:psnr_comparison}. We then refined the selected features using annotated masks from our dataset to focus on regions of interest, improving detection precision. Figure \ref{fig:haarfilter} illustrates the complete process, including Haar-like filter application, feature extraction, and mask-based refinement. This approach ensures that the Haar-like features in SHARP-Net are optimized for culvert and sewer pipe defect detection, potentially enhancing the model's performance and generalizability.

\begin{table}[!t]
\caption{PSNR comparison between the constructed images of different sliding window sizes for 1000 samples from Culvert-Sewer defects dataset.}
    \label{tab:psnr_comparison}
\centering
\begin{tabular}{|c||c|}
\hline
\textbf{Window Size} & \textbf{PSNR} \\
    \hline
        PSNR between (4,4) and (4,2) (4,2) & 4.5533 \\
        \hline
        PSNR between (16,4) and (4,2) (4,2) & 4.4030 \\
        \hline
        PSNR between (4,4) and (16,4) & 4.5284 \\
        \hline
        PSNR between (4,4) and (8,2) & 4.7518 \\
        \hline
        PSNR between (16,4) and (8,2) & 4.7397 \\
        \hline
        PSNR between (4,2) (4,2) and (8,4) (8,4) & 20.9121 \\
\hline
\end{tabular}
\end{table}

SHARP-Net incorporates Haar-like features through a feature injection gate in its second layer. This gate aligns extracted features with existing layers by creating matching convolutional layers and applying $1\times1$ convolutions for element-wise multiplication with input data. Figure \ref{fig:model} illustrates this process. Our research shows that Haar-like features significantly improve deep learning models' performance in semantic segmentation, particularly for infrastructure analysis like culverts and sewer pipes. Systematic exploration of feature extraction strategies has led to enhanced model performance and segmentation accuracy. SHARP-Net's unique architecture and state-of-the-art performance make it highly effective for complex tasks, including semantic segmentation of infrastructure and satellite imagery. This capability is crucial for accurate infrastructure management and monitoring. The successful integration of Haar-like features not only improves SHARP-Net but also demonstrates the potential for advancing semantic segmentation across various domains.


\section{Datasets}
This section is divided into two parts. Section \ref{sec:datasetCreation} details the development and characteristics of the Culvert-Sewer Defects dataset, created for this study. Section \ref{sec:DeepGlobedataset} introduces the benchmark DeepGlobe Land Cover Classification Dataset, used to evaluate SHARP-Net's performance and generalizability.

\subsection{Culvert-Sewer Defects  Dataset} \label{sec:datasetCreation}
In this subsection, we discuss the process of collecting inspection videos, converting them into frames, and performing pixel-wise annotation to create our 5000-image dataset.

\subsubsection{Data Collection and Preprocessing} We curated a comprehensive dataset comprising 580 annotated underground infrastructure inspection videos from two distinct sources. These videos encompass a wide range of real-world conditions encountered in both culverts and sewer pipes. The diversity in our dataset is substantial, capturing variations in materials, shapes, dimensions, and imaging environments that are typical of inspection scenarios. This ensures that our dataset is representative of the challenges faced during actual inspections.

Each video is accompanied by a detailed report prepared by skilled technicians. These reports document the types and locations of deficiencies observed throughout the inspections. 

To facilitate detailed analysis, we partitioned the videos into discrete frames, selecting intervals ranging from 4 to 10 seconds. This segmentation yielded approximately 5970 frames, with each frame corresponding to a specific deficiency as described in the accompanying report. Each image is time-stamped to the exact second, allowing for precise identification of the deficiency's location within the pipe according to the report. The resulting dataset, though extensive, presents a significant challenge due to the imbalanced distribution of deficiencies. Certain classes of deficiencies are significantly underrepresented, posing a challenge for model training and evaluation. This imbalance reflects the natural occurrence of various deficiencies in real-world inspection scenarios, adding another layer of complexity to the dataset.

\subsubsection{Pixel-wise Annotation for Semantic Segmentation} To prepare the dataset for training semantic segmentation models, it was essential to understand the specific requirements of the task. We opted for semantic segmentation over other methods, such as object detection or classification, due to its pixel-level precision. This precision allows for the identification of defects and features across the entire image, providing detailed spatial information. Consequently, semantic segmentation facilitates a comprehensive analysis of culverts and sewer pipes, ensuring that every part of the infrastructure is inspected thoroughly.

To achieve fine-grained semantic segmentation, skilled annotators manually outlined each deficiency instance within the video frames, generating precise pixel-level masks to serve as ground truth. This level of detail is crucial for accurately identifying and categorizing deficiencies at the pixel level, facilitating the development and evaluation of robust segmentation algorithms.
    
We categorized the semantic segmentation masks into nine common structural deficiency classes. The dataset exhibits significant class imbalance, with some classes being much more prevalent than others. The corresponding Class Importance Weights (CIW) are detailed in Table \ref{tab:CIW}. We employed the LabelMe tool to annotate the extracted video frames, forming our Culvert-Sewer Defects dataset.

\begin{table}[!t]
\caption{Class Importance Weights (CIW)}
\label{tab:CIW}
\centering
\begin{tabular}{|c||c|}
\hline
\textbf{Class} & \textbf{Importance Weight} \\
\hline
Water Level & 0.0310 \\
\hline
Cracks & 1.0000 \\
\hline
Roots & 1.0000 \\
\hline
Holes & 1.0000 \\
\hline
Joint Problems & 0.6419 \\
\hline
Deformation & 0.1622 \\
\hline
Fracture & 0.5100 \\
\hline
Encrustation/Deposits & 0.3518 \\
\hline
Loose Gasket & 0.5419 \\
\hline

\end{tabular}
\end{table}

Each annotated class is color-coded according to the US NASSCO's pipeline assessment certification program (PACP) guidelines \cite{pottawatomie}. A professional civil engineer assigned importance weights to each deficiency class based on their economic and safety impacts, which were normalized to prioritize learning during model training and used for Frequency-Weighted IoU (FWIoU).

\subsection{DeepGlobe Land Cover Classification  Dataset}\label{sec:DeepGlobedataset}


The satellite image benchmark datasets used in this study are from the DeepGlobe challenge \cite{demir2018deepglobe}. The datasets are derived from the DigitalGlobe Vivid+ collection, which focuses on rural areas. It includes seven land cover classes: agriculture land, urban land, rangeland, water, barren land, forest land, and unknown. Urban land consists of built-up areas with human artifacts; agriculture land includes farms, croplands, orchards, vineyards, and horticulture zones; rangeland is non-forest, non-farm green spaces and grasslands; forest land has at least 20\% tree crown density with clear cuts; water covers rivers, oceans, lakes, wetlands, and ponds; barren land includes mountains, rocks, deserts, beaches, and vegetation-free zones; and unknown areas are obscured by clouds or unclassifiable. The dataset has an online leaderboard and test metrics evaluated on hold-out test images. It is divided into three subsets: 803 training images, 172 test images, and 171 validation images. The test and validation sets consist of unlabeled images, about 30\% of the dataset. For the comparative analysis, only the annotated training samples were used, further divided into three segments for comprehensive evaluation and model validation.

\section{Experimental Setup and Training Protocol}

In this section, we provide an overview of the methodologies and parameters utilized in the development and training of our model. We detail the evaluation metrics employed, the optimization strategies and loss functions applied, as well as other critical aspects of the implementation, including training procedures, hardware and software configurations. 

\begin{itemize}
    \item Metrics Used for Evaluation: To evaluate the performance of our model in the semantic segmentation task, we employed several metrics, including Intersection over Union (IoU), Frequency-Weighted IoU (FWIoU), F1-Score, Balanced Accuracy, and Matthews Correlation Coefficient (MCC).
    \item Optimization and Loss Functions: For the optimization of our model, we used the Adam optimizer with a learning rate of \(10^{-3}\). Adam is an adaptive learning rate optimization algorithm that has been shown to work well in practice for many deep learning models. It combines the advantages of two other popular optimizers: AdaGrad, which works well with sparse gradients, and RMSProp, which works well in online and non-stationary settings.
    For the loss function, we employed the categorical cross-entropy loss. This loss function is particularly suitable for multi-class classification problems, as it measures the performance of a classification model whose output is a probability value between 0 and 1, which necessitates the use of one-hot encoding for our labels. The categorical cross-entropy loss calculates the difference between the true label and the predicted probability distribution, penalizing the model more heavily for larger errors. This helps guide the model to make more accurate predictions.
    
    \item Training Procedures: All models are trained for 100 epochs on Culvert-Sewer Defects dataset and 200 epochs on Deep Globe Land Cover benchmark dataset. The datasets are divided into three subsets: training (70\%), validation (15\%), and test (15\%). This split ensures that the models are evaluated on unseen data to assess their generalization performance. Additionally, baseline models are established and evaluated under the same experimental conditions to provide a comparison benchmark for assessing the performance of the proposed model.
    
     \item  Hardware and Software: All models are trained using NVIDIA T4 GPUs using Keras TensorFlow that facilitated the implementation and training of our model.
 
\end{itemize}

\section{Results and discussion}\label{sec:results_discuss}

This section evaluates our proposed model's performance against leading baseline and state-of-the-art semantic segmentation architectures. We analyze various metrics and organize our findings into the following subsections:

\subsection{Comparison with Baseline Architectures}
We compared our base SHARP-Net (without Haar-like features) to several models, as shown in Table \ref{tab:Performance}. The FPN model in our comparison uses a ResNet backbone pretrained on ImageNet and fine-tuned on our dataset to adapt to its specific characteristics.
\setlength{\parskip}{1em} 

We tested SegFormer-b0 and SegFormer-b5 models, both with and without ImageNet pretraining. This approach, also applied to the original FPN, allowed us to assess the impact of pretraining on model performance across different architectures.

\begin{figure} [ht]
    \centering
    \includegraphics[width=1\linewidth]{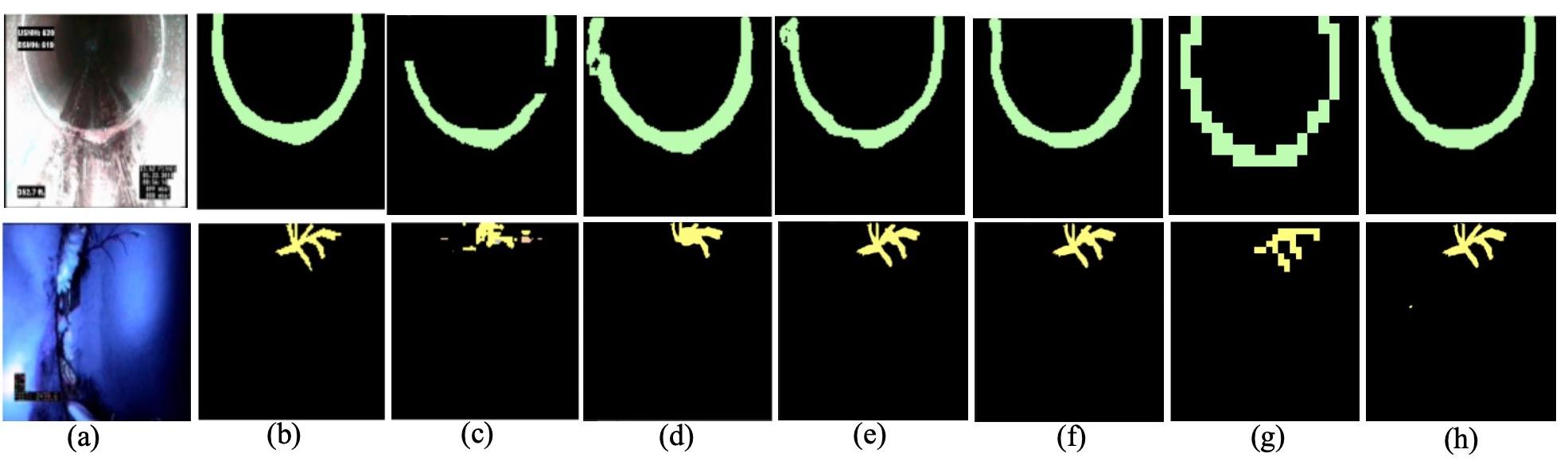}
  \caption{Comparative segmentation results on the culvert-sewer defects dataset are shown, with the first row illustrating joint problem defects and the second row depicting tree root problems: (a) Original images, (b) Ground truth, (c) U-Net, (d) CBAM U-Net, (e) FPN with ResNet, (f) ASCU-Net, (g) SegFormer, (h) The proposed models.}\label{fig:VComp-label}

\end{figure}

%
\begin{table}[!t]
\caption{Performance Comparison of Various Models on Culvert-Sewer Defects Dataset. w/bg: with background, w/o bg: without background}
\label{tab:Performance}
\resizebox{0.95\columnwidth}{!}{%
\centering
\begin{tabular}{|c||c||c||c||c||c||c|}
\hline
\textbf{Model} & \textbf{IOU w/ bg} & \textbf{IOU w/o bg} & \textbf{FWIoU} & \textbf{F1} & \textbf{Bal. Acc} & \textbf{MCC} \\

FPN with ResNet (original) & 0.69947 & 0.66575 & 0.69657 & 0.80610 & 0.81387 & 0.83922 \\
\hline
U-Net & 0.58559 & 0.53906 & 0.48980 & 0.69333 & 0.63078 & 0.40457 \\
\hline
CBAM U-Net & 0.60501 & 0.55889 & 0.67053 & 0.71269 & 0.67964 & 0.71296 \\
\hline
ASCU-Net & 0.70358 & 0.67021 & 0.71491 & 0.81161 & 0.79463 & 0.79451 \\
\hline
SefFormer MiT-B0  &0.56676 &  0.51632 & 0.59479 &0.70084 &0.67128 &0.69272\\
\hline
SefFormer MiT-B5 (1) &0.64326 &  0.60174 & 0.64025 &0.742488 &0.729208 &0.72586\\
\hline
SefFormer MiT-B5 (2) &0.59248 &  0.55087 & 0.61470 &0.70182 &0.69086 &0.72204\\
\hline
SHARP-Net base model & 0.77187 & 0.74601 & 0.78073 & 0.86346 & 0.84264 & 0.85005 \\
\hline

\end{tabular}
}
\end{table}
Figure \ref{fig:VComp-label} shows a visual analysis of the models evaluated in our study. This analysis is crucial for understanding the models' performance in semantic segmentation tasks, particularly in capturing fine-grained details.

U-Net and CBAM U-Net have limitations in reconstructing images despite accurately identifying the root. This is due to the architecture's struggle to preserve fine-grained spatial information through the encoder-decoder pathway. This leads to an 18.63\% decrease in Intersection over Union (IoU) scores compared to our proposed approach, especially in high-resolution feature preservation areas.

The SegFormer models (SegFormer-b0 and SegFormer-b5) consistently show visual artifacts in their output images. This is due to the use of dense layers in the decoder section. Our results suggest a trade-off in fine detail preservation, despite excelling in capturing global context. Quantitatively, our model has a 20.51\% higher IoU score compared to the SegFormer MiT-B0 models.

Our model excels in root image reconstruction, capturing fine root details for accurate and visually coherent reconstructions. This improvement is attributed to our model's feature pyramid network (FPN) architecture, which combines multi-scale feature representations [Cite]. The integration of Haar-like features enhances the model's edge and texture detection for accurate root segmentation.

Our model achieves a 7.24\% improvement in IoU over the next best performing model (ASCU-Net). Ablation studies show that incorporating Haar-like features contributes to a 22.74\% increase in IoU, demonstrating a significant improvement in segmentation accuracy.

These results highlight the effectiveness of our proposed architecture in handling complex image reconstruction tasks, especially those requiring the preservation of intricate details. The superior performance is visually and quantitatively significant, demonstrating the robustness of our approach across various segmentation quality metrics.

We evaluated our proposed model on the DeepGlobe land cover classification dataset, comparing it to two baseline architectures: the original U-Net and FPN. As shown in Table \ref{tab:PerformanceDeepGlob}, our model consistently outperforms both baselines across various metrics. Notably, it achieves an average IoU improvement of 10.7\% compared to U-Net and FPN. These results demonstrate our model's effectiveness and its ability to generalize across different datasets.

\begin{table}[!t]
\caption{Performance Comparison of Various Models on DeepGlobe Land Cover Classification Dataset. w/bg: with background, w/o bg: without background.}
\label{tab:PerformanceDeepGlob}
\resizebox{1\columnwidth}{!}{%
\centering
\begin{tabular}{|c||c||c||c||c||c|}
\hline
\textbf{Model} & \textbf{IOU w/ bg} & \textbf{IOU w/o bg}  & \textbf{F1} & \textbf{Bal. Acc} & \textbf{MCC} \\

\hline
FPN with ResNet (original) & 0.56090 & 0.55480&	0.69123&	0.58348&	0.64471 \\
\hline
U-Net & 0.61063 &	0.59214	&0.72956	&0.69600	&0.70181\\
\hline
SHARP-Net base model (this paper) &0.70641&	0.69899&	0.81641&	0.79231&	0.79798 \\
\hline
\end{tabular}
}
\end{table}

\begin{figure} [ht]
    \centering
    \includegraphics[width=0.99\linewidth]{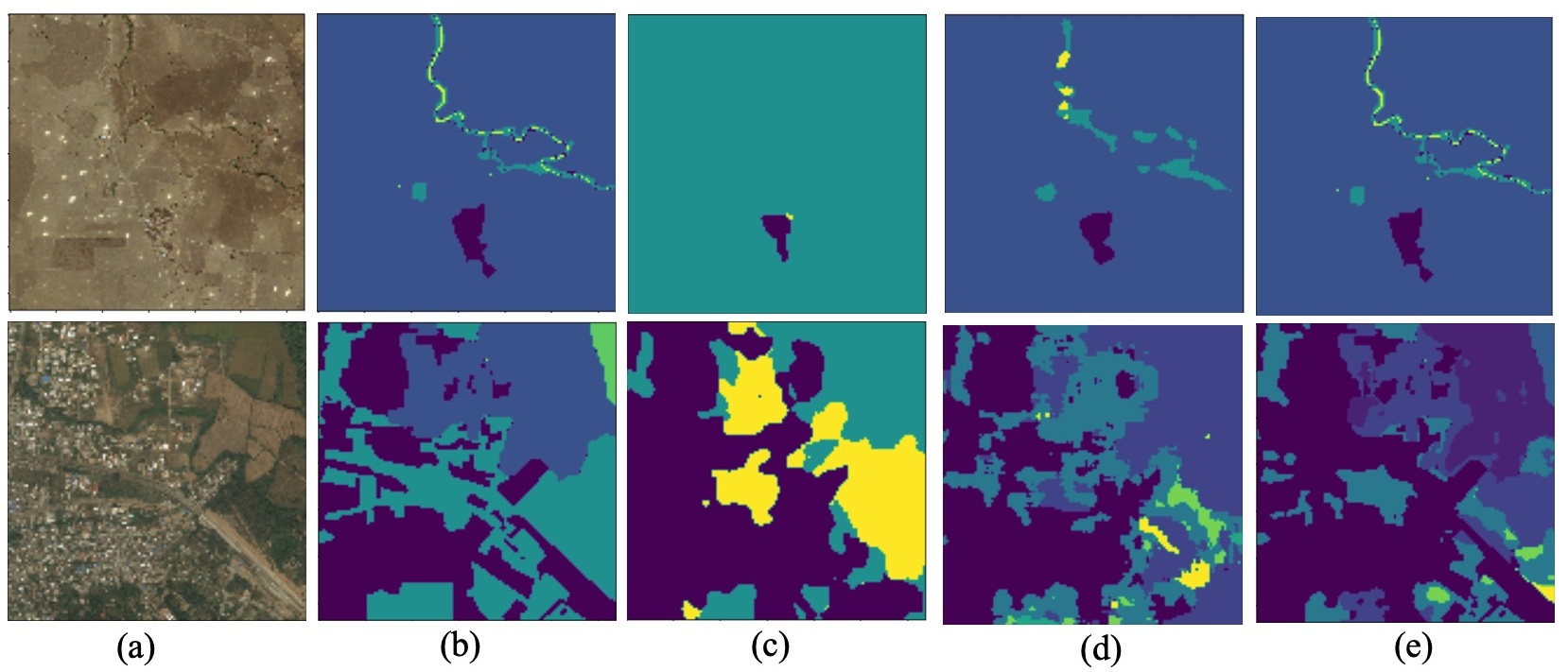}
  \caption{Comparative segmentation results on the DeepGlobe Land Cover Classification Dataset, with two samples showing different types of land cover: (a) Original images, (b) Ground truth, (c) FPN with ResNet, (d) U-Net, (e) The proposed model.}\label{fig:deepGlobe}

\end{figure}

Figure \ref{fig:deepGlobe} show that while it is evident that all models could benefit from further refinement to improve their accuracy, our proposed model consistently outperforms the baseline models. This superior performance is demonstrated across various test cases within the dataset, highlighting our model’s enhanced capability in accurately classifying and segmenting different land cover types. This visual comparison underscores the robustness and effectiveness of our model in handling complex segmentation tasks compared to the original U-Net and FPN models.

\subsection{Model Efficiency and Computational Performance}
The proposed model is remarkably efficient, with only 1.32 million parameters, representing a 19-24 times reduction compared to baseline models (Table \ref{tab:numberofparams}). This reduction has important implications for model performance and applicability.

The dramatic decrease in parameter count reduces computational complexity, crucial in resource-constrained environments like edge devices or mobile platforms with limited computational power and memory. Our lean architecture enables faster inference times and lower memory footprint, making it suitable for real-time applications in fields like autonomous vehicles, mobile health diagnostics, or on-site infrastructure inspection.

The reduced parameter count reduces the risk of overfitting on smaller datasets. With fewer parameters, the model is less likely to memorize training data and more likely to generalize well to unseen examples. This is valuable in domains with scarce or expensive large annotated datasets, such as specialized medical imaging or rare defect detection in industrial applications.

The model’s efficiency affects training time and energy consumption. It needs less resources and time to train with fewer parameters, potentially reducing the carbon footprint of model development. This aligns with the emphasis on sustainable AI and green computing in the ML community.

Our model maintains competitive performance, despite the reduced parameters. This suggests that the architecture efficiently captures essential task features, eliminating redundant or less informative parameters. Achieving high performance with fewer parameters underscores the effectiveness of our design choices, including depth-wise separable convolutions and Haar-like features.

Our proposed model's efficiency offers benefits in computational performance, generalization ability, and practical applicability across various scenarios with computational constraints, characterized by its reduced parameter count. This efficiency, coupled with the model's performance, positions it as a valuable contribution to semantic segmentation, especially for applications requiring accuracy and resource utilization balance.

\begin{table}[!t]
\caption{Comparison of the Number of Trainable Parameters in Different Models}  \label{tab:numberofparams}%
\centering
\begin{tabular}{|c||c|}
\hline
    \textbf{Model} & \textbf{Number of Trainable Parameters} \\ 
    \hline
    U-Net & 31,032,521 \\
    \hline
    CBAM U-Net & 31,221,065 \\
    \hline
    ASCU-Net & 31,841,202 \\
    \hline
    FPN with ResNet (original) & 25,698,557 \\
    \hline
    SegFormer MiT-b5 & 84,601,801 \\
    \hline
    SegFormer MiT-b0 & 3,716,457 \\
    \hline
    SHARP-Net (this paper) &   1,324,660    \\
    \hline
\end{tabular}
\end{table}

\subsection{Impact of Haar-Like Features}
This section explores how Haar-like features enhance deep learning models, particularly SHARP-Net. Table \ref{tab:model_performance} shows the performance improvements achieved by gradually adding Haar-like features to SHARP-Net. Starting with the base model, we systematically incorporated additional features to measure their individual and combined effects on model performance.

%
\begin{table}[!t]
  \caption{Performance metrics comparison for HFFPN models using feature engineering.}
    \label{tab:model_performance}
    \resizebox{1.0\columnwidth}{!}{%
\centering
\begin{tabular}{|c||c||c||c||c||c||c|}
\hline
\textbf{Model} & {\textbf{IOU w/ bg}} & {\textbf{IOU w/o bg}} & {\textbf{FWIoU}} & {\textbf{F1}} & {\textbf{Bal. Acc}} & {\textbf{MCC}} \\
        SHARP-Net (base model) & 0.77187 & 0.74601 & 0.78073 & 0.86346 & 0.84264 & 0.85005 \\
        \hline
        SHARP-Net using two edge features & 0.89413 & 0.88136 & 0.89705 & 0.94191 & 0.93004 & 0.93985 \\
        \hline
        SHARP-Net using the first three Haar-Like features & 0.94750 & 0.94109 & 0.96208 & 0.97223 & 0.96313 & 0.97177 \\
        \hline
        SHARP-Net using five Haar-Like features & 0.94752 & 0.94110 & 0.96689 & 0.97831 & 0.97351 & 0.98056 \\
        \hline
\end{tabular}
}
\end{table}

We applied our Haar-like feature integration technique to the U-Net architecture, demonstrating its versatility across different deep learning models. As shown in Table \ref{tab:UPerformance}, this integration improved performance by 35.01\% compared to the original U-Net. This significant enhancement highlights the potential of Haar-like features to boost various semantic segmentation models beyond SHARP-Net.
%
\begin{table}[!t]
\caption{Performance Comparison of U-Net Architecture with and without Haar-like Features on the Culvert-Sewer Defects Dataset. w/bg: with background, w/o bg: without background.}
\label{tab:UPerformance}
\resizebox{1.0\columnwidth}{!}{%
\begin{tabular}{|c||c||c||c||c||c||c|}
\hline

\textbf{Model} & \textbf{IOU w/ bg} & \textbf{IOU w/o bg} & \textbf{FWIoU} & \textbf{F1} & \textbf{Bal. Acc} & \textbf{MCC} \\
\hline
U-Net & 0.58559 & 0.53906 & 0.48980 & 0.69333 & 0.63078 & 0.40457 \\
\hline
U-Net using five Haar-Like features & 0.79063 &   0.77428  &   0.84016 &  0.84516 & 0.86452     &  0.80270 \\
\hline
\end{tabular}
}
\end{table}

\begin{figure}[ht]
    \centering
    \includegraphics[width=1.0\linewidth]{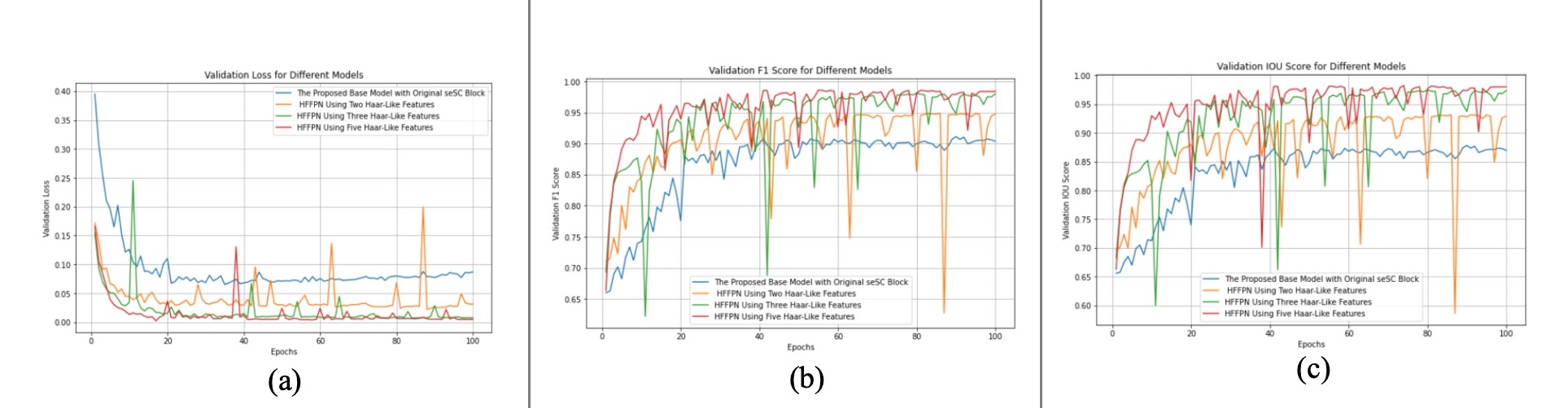}
    \caption{Validation graphs comparing the SHARP-Net (red) to other models on the sewer-culvert defects dataset: (a) loss, (b) F1-score, and (c) IoU. The performance gap between the base model and models with injected features is evident. Additionally, the performance improves as more features are incorporated into the model, which also enhances training stability.}
    \label{fig:Validation-label}
\end{figure}

The results confirm that incorporating Haar-like features significantly improves performance. While using three features produces results similar to five features, adding more features speeds up convergence and reduces training time. Peak performance was reached within 20 epochs, and using five features improved training stability (see Figure \ref{fig:Validation-label}). These findings demonstrate that Haar-like features enhance both model accuracy and training efficiency.


\begin{figure}
    \centering
    \includegraphics[width=0.99\linewidth]{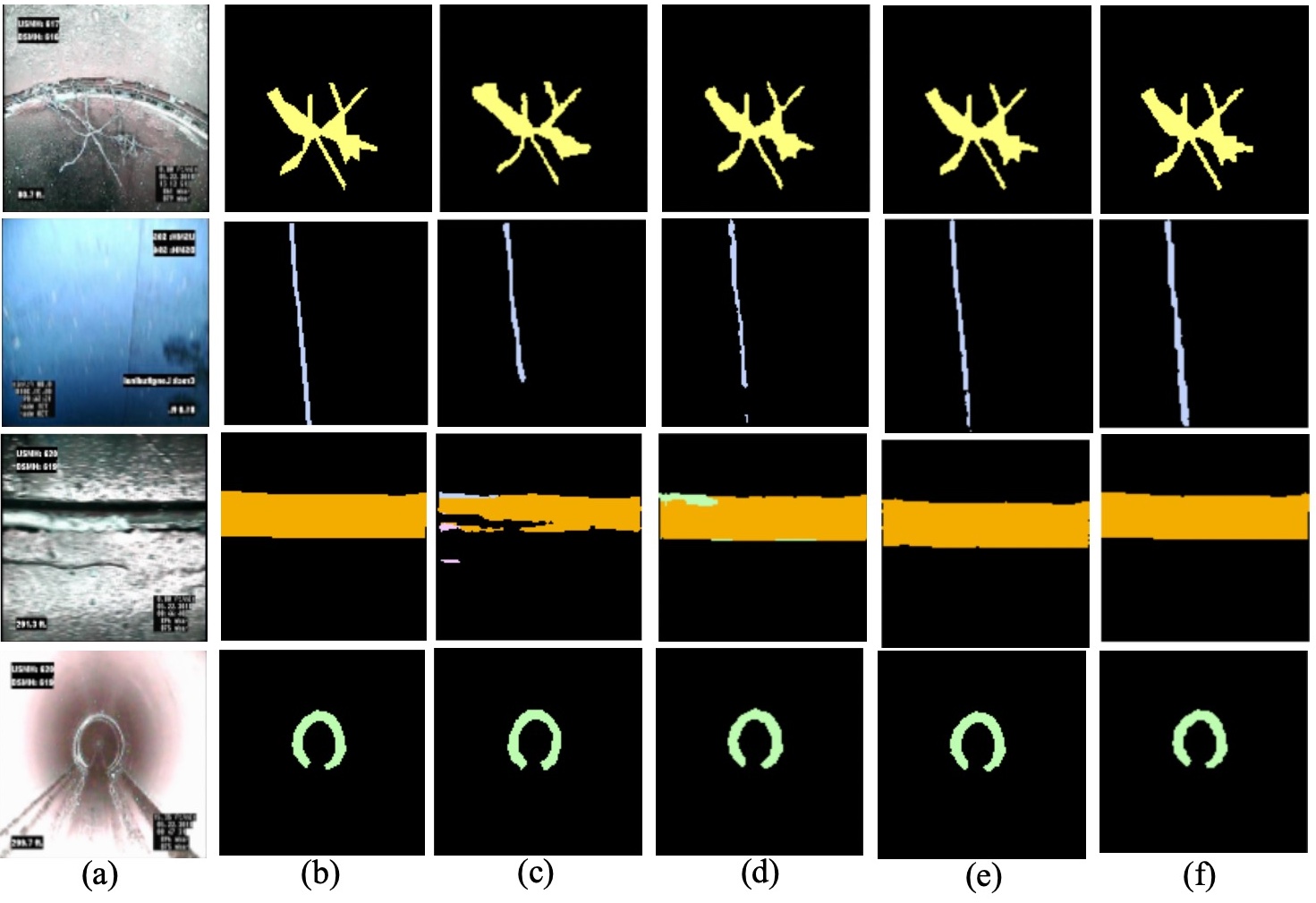}
    \caption{Visual comparisons of SHARP-Net results with varying Haar-like features on sewer-culvert defects: (a) Original images, (b) Ground truth, (c) Base model, (d) Two edge features, (e) Three Haar-like features, and (f) Five Haar-like features yielding the highest quality reconstructions.}
    \label{fig:enter-label}
\end{figure}


Figure \ref{fig:enter-label} provides a qualitative analysis of segmentation results from various SHARP-Net configurations. The first and last rows show consistently well-segmented samples across all model variants, demonstrating the base architecture's robust performance. The second and third rows highlight improvements achieved through Haar-like features, showcasing the model's enhanced ability to capture fine details and complex structures. Figure \ref{fig:Validation-label} presents quantitative evidence supporting the effectiveness of Haar-like features. SHARP-Net exhibits faster convergence and lower validation loss (Fig. \ref{fig:Validation-label}a), aligning with reduced training time observations. F1-Score (Fig. \ref{fig:Validation-label}b) and IoU scores (Fig. \ref{fig:Validation-label}c) further demonstrate SHARP-Net's superior performance across multiple metrics. In short, SHARP-Net consistently outperforms baseline architectures on both the Culvert-Sewer Defects and DeepGlobe Land Cover datasets. The integration of Haar-like features yields a 20-30\% improvement in IoU scores, highlighting the significant benefits of this approach. The method's versatility is evident in its successful application across diverse models and datasets, demonstrating its potential to advance semantic segmentation across various domains.


\section{Conclusion}
We present SHARP-Net, a novel deep learning architecture for precise semantic segmentation on challenging multiclass datasets. SHARP-Net combines a bottom-up top-down structure with sparsely connected blocks, depth-wise separable convolutions, and Haar-like feature extraction. This design addresses issues like irregular defect shapes, occlusions, limited data, and class imbalance. Our evaluation on the Culvert-Sewer Defects and DeepGlobe Land Cover Classification datasets shows SHARP-Net's superior performance. The base model achieved IoU scores of 77.2\% and 70.6\% on these datasets, respectively, with only 1.32 million parameters. Adding Haar-like features improved IoU to 94.75\%, outperforming state-of-the-art architectures like FPN, U-Net, CBAM U-Net, SegFormer, and ASCU-Net. Haar-like features not only enhanced accuracy but also accelerated convergence, reducing training time and computational requirements. This technique can potentially improve other deep learning models by at least 20\%.

While SHARP-Net performs well, its efficacy across diverse semantic segmentation tasks and high-resolution or real-time applications, and feature selection optimization needs further investigation. Future research should focus on automating feature selection, exploring cross-domain adaptability, and optimizing for edge deployment. Additionally, incorporating temporal consistency for video segmentation, integrating multimodal data, and enhancing model interpretability will be crucial. These advancements aim to broaden SHARP-Net's applicability and push the boundaries of semantic segmentation.

\section*{Acknowledgement}
This research was partly supported by the U.S. Department of the Army – U.S. Army Corps of Engineers (USACE) under contract W912HZ-23-2-0004. The views expressed in this paper are solely those of the authors and do not necessarily reflect the views of USACE.

\bibliographystyle{IEEEtran}
\bibliography{references}

\appendix
\section{Supplementary Material} \label {sec:SupplementaryMaterial}

This section details the experiments performed to extract and refine the Haar-like features.

\subsection{Haar-like Feature Experiments} \label{sec:HaarAnalysis}

This section elaborates on the experiments performed to refine the Haar-like feature extraction and integration within our proposed model. Detailed results and analyses of various Haar-like feature configurations are provided below.

\begin{figure}
    \centering
    \includegraphics[width=0.99\linewidth]{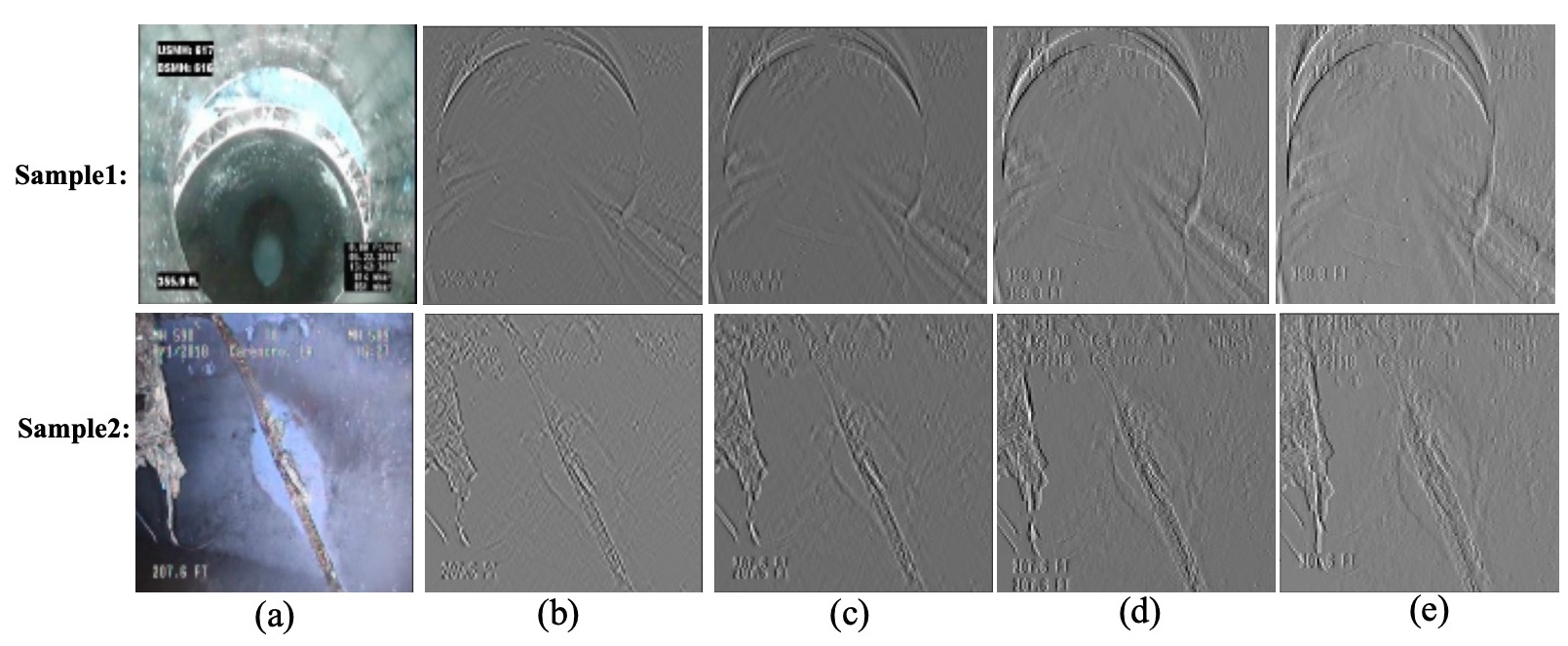}
    \caption{Comparative results of Haar-like features using different window sizes on two samples from the Culvert-Sewer Defects dataset: (a) Original images, (b) Filter response for window size (2,2), (c) Filter response for window size (4,2), (d) Filter response for window size (6,2), and (e) Filter response for window size (8,2).}
    \label{fig:haarsup1}
\end{figure}

\paragraph{Window Size Variations} We evaluated Haar-like features with various window sizes to determine their impact on feature extraction. Starting with smaller window sizes, such as (2,2), we observed that this configuration was effective for detecting fine details but had limitations in capturing broader contextual information. Increasing the window size to (4,2) enhanced feature extraction by providing more detailed analysis, though it also led to increased computational requirements.
Further experimentation with larger window sizes, such as (8,2) and (16,2), revealed that while these sizes improved edge detection sharpness, they also introduced pixel artifacts and continued to demand significant computational resources. Figure \ref{fig:haarsup1} shows a visual comparison of these windows and their filter responses on two samples from the Culvert-Sewer Defects dataset. The Peak Signal-to-Noise Ratio (PSNR) was used to quantify the effectiveness of these window sizes, with the results summarized in Table \ref{tab:psnr_sample_comparison}. The PSNR values, although relatively low, provided insights into the quality of the reconstructed features and guided the selection of optimal window sizes. 

\begin{table}[h]
    \centering
    \caption{PSNR comparison between the real and constructed image of four different sliding window sizes for Sample 1 and 2 from the Culvert-Sewer Defects dataset.}
    \label{tab:psnr_sample_comparison}
    \begin{tabular}{lrr}
        \toprule
        \textbf{Window Size} & \textbf{PSNR-Sample 1} & \textbf{PSNR-Sample 2} \\
        \midrule
        Size (2,2) & 5.6256 & 4.7701 \\
        Size (4,2) & 5.6102 & 4.7551 \\
        Size (8,2) & 5.6043 & 4.7529 \\
        Size (16,2) & 5.6072 & 4.7518 \\
        \bottomrule
    \end{tabular}
\end{table}

\begin{figure}
     \centering
     \includegraphics[width=0.99\linewidth]{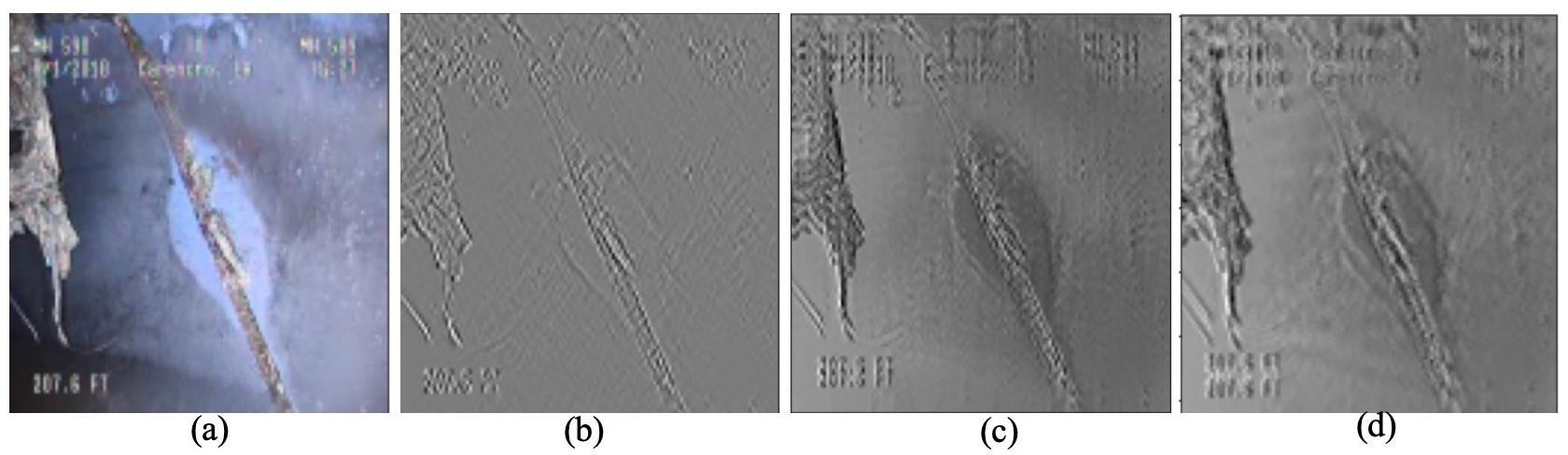}
     \caption{Comparative results of single versus doubled size filters on a sample from the Culvert-Sewer Defects dataset: (a) Original image, (b) Filter response for window size (4,2), (c) Filter response for window size (4,2) × (4,2), and (d) Filter response for window size (8,4) × (8,4).}
     \label{fig:haarsup2}
 \end{figure}

We also tested various Haar-like filters, including both rectangular and diagonal configurations. Filters such as (4,4), (8,4), and (16,4) were used to capture different types of features, including edges and lines. The performance of these filters was also evaluated based on PSNR values and visual inspection. Additionally, we experimented with double-sized filters, such as (4,2)(4,2) and (8,4)(8,4), which yielded clear filter responses as illustrated in Figure \ref{fig:haarsup2}. Overall, the results demonstrated that different Haar-like filters provided varying degrees of feature detection effectiveness, indicating that incorporating a diverse set of Haar features can significantly enhance model training.

Following these experiments, we aimed to include only the features that were distinct from each other to avoid redundancy and increase diversity. Therefore, the window sizes (4,2), (4,4), (8,4), and (16,4) were selected for further use based on their PSNR values as detailed in the main text.

\paragraph{Feature Refinement and Integration} The extracted Haar-like features were refined using annotated masks from the dataset, focusing on regions of interest to improve feature quality. This is done by multiplying the manually annotated mask by the filter responses. These refined features were integrated into the model through a feature injection gate, as discussed in the main text, aligning them with the model's layers to enhance segmentation performance.

In summary, the effectiveness of various Haar-like features was compared using PSNR values and visual inspection. Features with higher PSNR values were selected for integration, while those with lower values were excluded to avoid redundancy and ensure the inclusion of the most informative features. Our experiments demonstrated that Haar-like features significantly improve model performance, particularly for tasks requiring fine-grained analysis. The integration of these features into the SHARP-Net model led to substantial performance gains, highlighting the versatility and effectiveness of Haar-like features in semantic segmentation.

\vfill

\end{document}